# LGENet: Local and Global Encoder Network for Semantic Segmentation of Airborne Laser Scanning Point Clouds

Yaping Lin[1], George Vosselman[1], Yanpeng Cao[2], Michael Ying Yang[1]

[1] Dept. of Earth Observation Science, Faculty ITC, University of Twente, Enschede, The Netherlands - (y.lin, george.vosselman, michael.yang)@utwente.nl

[2] State Key Laboratory of Fluid Power and Mechatronic Systems, School of Mechanical Engineering, Zhejiang University, Hangzhou, China - caoyp@zju.edu.cn

## Abstract

Interpretation of Airborne Laser Scanning (ALS) point clouds is a critical procedure for producing various geo-information products like 3D city models, digital terrain models and land use maps. In this paper, we present a local and global encoder network (LGENet) for semantic segmentation of ALS point clouds. Adapting the KPConv network, we first extract features by both 2D and 3D point convolutions to allow the network to learn more representative local geometry. Then global encoders are used in the network to exploit contextual information at the object and point level. We design a segment-based Edge Conditioned Convolution to encode the global context between segments. We apply a spatial-channel attention module at the end of the network, which not only captures the global interdependencies between points but also models interactions between channels. We evaluate our method on two ALS datasets namely, the ISPRS benchmark dataset and DCF2019 dataset. For the ISPRS benchmark dataset, our model achieves state-of-the-art results with an overall accuracy of 0.845 and an average F1 score of 0.737. With regards to the DFC2019 dataset, our proposed network achieves an overall accuracy of 0.984 and an average F1 score of 0.834.

## 1 Introduction

With the advanced techniques of light detection and ranging (LiDAR) systems, point clouds are more easily obtained in various scenes. Airborne laser scanning (ALS) point clouds have become an essential type of data in the generation processes of digital terrain models (DTM) (Chen et al., 2017), landscape models (Murtha et al., 2018), 3D city models (Lin et al., 2018) and land use maps (Meng et al., 2012). These point cloud based products are required in many disciplines, like urban planning (Murgante et al., 2009), land administration (Lemmen et al., 2015), forest inventory (Wallace et al., 2012), tourism (Cooper et al., 2013) and disaster management (Shen et al., 2010). The interpretation of ALS point clouds is a prerequisite for their use in these applications. One of the interpretation methods is semantic segmentation which assigns a semantic label to each point in the dataset. Manually labelling every point is quite time-consuming, especially for large urban areas. Thus, machine learning techniques are developed to automate the interpretation process (Vosselman and Maas, 2010).

Machine learning approaches used for 3D scene understanding traditionally focused on extracting representative handcrafted features to describe local geometry (Lin et al., 2014; Weinmann et al., 2013) and training different discriminative classifiers to produce pointwise labels like Supported Vector Machine (SVM) (Lodha et al., 2006), AdaBoost (Lodha et al., 2007), random forests (RF) (Chehata et al., 2009), Gaussian Mixture Model (GMM) (Weinmann et al., 2014) and Artificial Neural Networks (ANN) (Xu et al., 2014). The involvement of contextual information between points has been proven to be effective in improving semantic segmentation results and this can be achieved by using graphical models such as Conditional Random Field (CRF) (Niemeyer et al., 2016; Vosselman et al., 2017). However, in these methods, low dimensional handcrafted features are not representative to distinguish all categories in the dataset especially for the ALS point clouds acquired over complicated scenes where objects are largely different in size.

Recently, deep learning methods have shown their powerful abilities in object recognition and semantic segmentation from images. The huge success of deep learning is due to learning features from different levels based on data instead of using the predefined features in traditional machine learning methods. Inspired by the success of deep learning in image related tasks, many deep learning based approaches for 3D interpretation tasks are proposed, like image-based methods (Boulch et al., 2018; Kalogerakis et al., 2017), voxel-based methods (Maturana and Scherer, 2015; Tchapmi et al., 2017) and point-based methods (Li et al., 2018; Qi et al., 2017a; Thomas et al., 2019). With regards to ALS point clouds, some researchers convert point clouds into sets of images (Hu and Yuan, 2016; Yang et al., 2017; Zhao et al., 2018) or 3D voxel grids (Schmohl and Sörgel, 2019). Others design networks (Arief et al., 2019; Li et al., 2020a; Winiwarter et al., 2019; Yousefhussien et al., 2018) that can directly consume ALS point clouds and learn more representative features with less information loss.

Global contextual cues have proven that they can further improve the results from deep learning methods for computer vision tasks. Some approaches use fully connected CRF to enforce global consistency and refine semantic predictions on images (Zheng et al., 2015) and point clouds (Tchapmi et al., 2017). Motivated by the self-attention module proposed by Vaswani et al. (2017) for machine translation, various other approaches adapt this concept for computer vision tasks like semantic segmentation of images (Wang et al., 2018) and point clouds (Feng et al., 2020). Nevertheless, these methods explore dependencies between pixels or points, ignoring relationships between objects which are informative for large scale complex outdoor scenes. Notably, Super point graph (SPG) (Landrieu and Simonovsky, 2018) only assigns labels to segments and incorrect segmentation causes errors in the final pointwise predictions. Therefore, it is still challenging to make use of global context at both point and object levels for large scales ALS data.

In this paper, we propose a novel 3D convolutional network, a local and global encoder network (LGENet), that can embed more representative features for ALS data and exploit global context at both object and point levels. Considering that the variance of ALS point cloud coordinates is larger in the XY plane than along the Z-axis, we first enhance the representativeness of features obtained by 3D convolutions by adding 2D convolutions in order to pay more attention to the point distribution on the XY plane. Next, motivated by SPG (Landrieu and Simonovsky, 2018), we encode global interdependencies between segments by segment-based Edge Conditioned Convolution (SegECC). Segments are obtained from the unsupervised algorithms before the training and trainable edge conditioned convolutions are applied to capture the spatial dependencies between objects. This operation can be inserted after any convolutional layers in the network. Finally, a spatial-channel attention is introduced to semantic segmentation of point clouds and placed at the end of the network to capture long-range interactions between points and dependencies between channels. The major **contributions** of this paper are listed as follows:

- We propose a hybrid block that combines features extracted from both 2D and 3D convolutions. 2D convolutions are introduced to allow the network to learn representative features for point clouds primarily distributed in horizontal dimensions.
- To capture global spatial dependencies at the object level, we design a SegECC operation that constructs graphs on segments and exploits the relationships among objects. Segment features are then concatenated to pointwise features to allow the network to adaptively encode local-global features.
- To make use of spatial and channel dependencies, a spatial-channel attention is modified for semantic segmentation of ALS point clouds. The spatial attention learns the global interactions between points and channel-wise attention enhances the discriminability of learned features for different semantic categories.

The remainder of the paper is structured as follows. In Section 2, we review related traditional methods and recent deep learning methods in semantic segmentation of point clouds. Section 3 introduces the hybrid convolution, SegECC and spatial-channel attention designed in our network. In Section 4, we show our results on the ISPRS benchmark dataset (Niemeyer et al., 2014) and compare LGENet against with other state-of-the-art models. Extensive ablation experiments are carried out on the ISPRS benchmark dataset (Niemeyer et al., 2014) to evaluate our proposed method. We also test our model with the DFC2019 dataset (Bosch et al., 2018). Section 5 draws conclusions of this paper.

# 2 Related work

## 2.1 Traditional methods

Traditional machine learning methods for classifying point clouds are generally divided into two steps, extracting handcrafted features and training discriminative classifiers. For semantic segmentation of ALS datasets, many researchers define features that describe local geometry. Lin et al., 2014 propose a method to compute 'eigenfeatures' from the covariance matrix of local neighboring points to characterize local point distributions of ALS point clouds, e.g. planarity, sphericity, linearity. Then a Support Vector Machine (SVM) is used to classify point clouds. Weinmann et al. (2013) evaluate 21 geometric features including 8 'eigenfeatures' derived from optimal local neighbors and a set of 2D geometric features to describe the local characteristics. These features are then tested with four classifiers, namely, nearest neighbor, k Nearest Neighbor, Naive Bayesian and SVM. However, these methods take each point's local geometry independently for pointwise prediction and ignore the spatial dependencies, resulting in prediction noises and label inconsistency.

The above issues can be addressed by taking advantage of the contextual information. An important statistical method to model the context is probabilistic graphical models, such as Conditional Random Fields (CRFs). Niemeyer et al. (2014) propose a pointwise classification method using CRF for ALS datasets. Unary potential is the pointwise probability distribution over classes produced by a learned classifier. Pairwise potential, revealing prominent relations between the data and object classes, is also learned during the training. Although this CRF based method gives rise to smoother results and improves class-specific accuracy, especially for classes with fewer instances, the pairwise CRF still takes interactions at a very local level into account and cannot avoid incorrect labelling to isolated point clusters. A longer-range of interactions between points is a possible solution. Xiong et al. (2011) propose a sequence of stacked classification procedures. They propagate pointwise classification to segments and then consider contextual information according to the segment-based results for the final pointwise prediction. Niemeyer et al. (2016) propose a two-layer hierarchical higher order CRF for semantic segmentation of ALS data based on the robust $P^n$ Potts model (Kohli et al., 2009) . The first layer, operating on points, takes both handcrafted geometric features and relations between points to produce pointwise labelling. In the second layer, nodes are represented by segments which are generated by a variant of the region growing algorithm, so that interactions between objects are considered. However, these methods need to extract handcrafted features before the training which are not representative for multiple categories in point clouds acquired from complex scenes.

## 2.2 Deep learning methods

The effectiveness of deep learning approaches has been proven in recent research and the idea of deep learning has been applied to point clouds interpretation.

As CNNs are capable of learning highly representative features in many image processing tasks, many strategies are proposed to adjust classical 2D image deep neural networks to 3D point clouds. One branch of methods is based on the concept of converting the unordered and irregularly distributed point clouds into rasterized 2D representations which are the input of the CNNs. For example, Kalogerakis et al. (2017) propose a fully convolutional network, ShapePFCN, for 3D part segmentation. The network input is rendered images of 3D shapes captured from different views. In addition to ShapePFCN, this projection-based method has been extended to the semantic segmentation of large-scale point clouds with complicated scenes. Boulch et al. (2018) generate images containing geometric features obtained from depth and RGB information. Then fully convolutional networks produce pixel-wise labels and these labels are converted into 3D space through a fast back-projection. Nevertheless, self-occlusion is difficult to avoid during the projection, especially for complicated outdoor scenes. For semantic segmentation of large-scale ALS data, numerous methods convert 3D point clouds into 2D rasterized features from the top view in order to pass the data through image based CNNs. Hu and Yuan (2016) conduct the ground points labelling of ALS data by assigning simple attributes to each pixel like minimum, maximum and mean of the height within each grid cell. Similarly, Yang et al. (2017) also apply 2D grids to 3D point clouds but they assign more full-waveform and geometric features to each 2D grid. Zhao et al. (2018) produce multi-scale contextual images that represent point set features like height, intensity and roughness. These methods of processing ALS point clouds require complicated features to be produced before the network training. Many pre-calculated features can be redundant and require large memory for data processing during the training. Also, only extracting features from projected point clouds on two 2D space leads to information loss along the third dimension.

Volumetric approaches that voxelize unordered point clouds into regular 3D grids are alternatives to processing point clouds in order to adapt to deep neural networks. Maturana and Scherer (2015) convert the sparsely distributed point clouds into $32 \times 32 \times 32$ binary occupancy grids where each voxel is categorized into occupied and unoccupied. Then voxelized point clouds are processed by 3D convolutions for fast object detection. 3DShapeNet (Wu et al., 2015) also uses binary 3D voxel grids as the network input for object recognition and shape completion. SegCloud (Tchapmi et al., 2017) is a 3D CNN that generates coarse down-sampled labels for each voxel. Then pointwise labels are obtained by transferring the voxel labels back points through trilinear interpolation. Concerning ALS datasets, Schmohl and Sörgel (2019) take voxelized ALS point clouds as the input of sparse submanifold convolutional networks (SSCNs). The voxelization unavoidably leads to information loss and causes artifacts. These disadvantages negatively impact the learning of representative 3D features. In addition, a large amount of unoccupied grids stored in voxel structures result in high memory requirements.

Recent research focuses on how to make the deep neural network directly consume point clouds to minimize information loss. PointNet, a deep learning network designed by Qi et al.

(2017), can directly process unstructured points without any rasterization or voxelization and it achieves compelling performance on a series of point cloud related tasks, like object classification, part segmentation and semantic segmentation. PointNet learns representative point set features by Multilayer Perceptron (MLP) layers. Spatial transformers which produce transformation matrices are also auxiliary learned to align input point clouds to a canonical space and improve the robustness to geometric transformations. The key limitation is that PointNet treats each point independently. It can only encode each point individually and aggregate point features into one global representation, failing in capturing local structures. To address the above issues, Qi et al. (2017b) present a hierarchical deep network called PointNet++. It consists of a sequence of set abstract modules that progressively capture geometric features in wider and wider local regions.

As 2D convolutional kernels have shown their effectiveness in capturing relationships in local neighbourhoods, deep neural networks based on the concept of 3D convolutions are proposed to extract representative features from local structures of point clouds. Unlike Voxnet (Maturana and Scherer, 2015) which only takes a grid-style input, these networks are able to directly process irregularly distributed point clouds and some of them define convolutional function over continuous 3D space where weights of points within a local neighbourhood depend on their spatial distributions around the central point. For example, Kernel Point Convolutions (KPConv) proposed by Thomas et al. (2019) are defined over continuous space. Linear correlation between point positions and kernel point positions defines weights of points to different areas inside convolutional kernels. Kernel point positions are learnable and this helps convolution kernels to adapt to local structures in a better way. Similarly, InterpConv (Mao et al., 2019), PointConv (Wu et al., 2019), SpiderCNN (Xu et al., 2018), and Flex-Convolution (Groh et al., 2019) are 3D convolutional operators defined over continuous space to capture local contextual information.

With regards to semantic segmentation of ALS point clouds, Yousefhussien et al. (2018) modify the PointNet and make the network learn from more input features which consist of XYZ coordinates and corresponding radiometric features extracted from IR-R-G imagery. AlsNet based on PointNet++ is proposed by Winiwarter et al. (2019). A batching framework is introduced to allow the network to process large scale point clouds. Lin et al. (2020) also apply PointNet++ to large scale ALS data and an active and incremental learning strategy is proposed to make the training more efficient. An Atrous XCRF network is designed by Arief et al. (2019) to avoid the overfitting during deep network training (eg. PointCNN) for ALS datasets that are small in size. Wen et al. (2020) first project 3D points to a horizontal plane and then use a directionally constrained point convolution to encode neighbouring orientation information in ALS data. Li et al. (2020) apply a dense hierarchical architecture with geometry-aware convolutions and an elevation-attention module to fully embed characteristics of ALS point clouds.

Exploiting global contextual information is also researched in 3D deep neural networks for semantic segmentation of point clouds. Tchapmi et al. (2017) use a fully connected conditional random field (FC-CRF) at the end of the 3D CNN to exploit long-range interactions among points. The FC-CRF is implemented as a differentiable Recurrent Neural Network. This formulation allows the joint training of 3D CNN and 3D FC-CRF. Landrieu and Simonovsky (2018) first partition large point clouds into geometrical homogenous point sets called superpoints and then apply graph convolutions to the graph constructed by superpoints. Gated Recurrent Units are implemented to exploit the long-range relationships among superpoints. Huang et al. (2020) achieve global optimization for semantic segmentation of ALS point clouds through the Markov random fields algorithm which is a post-processing to refine initial classification results.

## 2.3 Attention models

Attention can be used as a tool to pay more attention to the most informative signals during the data processing and attention models have been widely used in natural language processing and computer vision tasks. Recently, the attention mechanism has shown its potential in encoding global contextual information. Vaswani et al. (2017) propose a self-attention module for machine translation. The idea is to encode the context at one position in a sequence by calculating a weighted average of embeddings at all positions. With regards to computer vision tasks, Wang et al. (2018) propose a non-local operation in order to capture long-range spatial dependencies. The non-local operation produces attention maps by computing the correlation between all possible point pairs in the feature space and those attention maps guide the aggregation of spatial contextual information. Apart from modelling spatial dependencies, channel-wise relationships are also exploited by attention mechanisms in order to enhance the representative power of deep learning models. For example, Hu et al. (2018) design squeeze-and-excitation blocks which firstly squeeze spatial features into a channel descriptor for each channel and then recalibrate channel-wise features by modelling channel-wise interdependencies. DANet (Fu et al., 2018) takes advantage of both spatial and channel-wise attentions. Their outputs are fused at the end of networks to boost feature representation, contributing to more precise predictions. Concerning the semantic point cloud segmentation, Feng et al. (2020) insert several pointwise spatial attention modules into deep neural networks to make use of interdependencies among all points regardless of their distance. The effectiveness of the pointwise spatial attention has been proven on the ShapeNet (Wu et al., 2015) and two indoor datasets namely, ScanNet (Dai et al., 2017) and S3DIS (Armeni et al., 2016). However, no outdoor dataset is tested in their experiments. Thus, it is valuable to explore how to take advantage of attention modules to boost feature representation of networks by modelling both spatial and channel-wise interdependencies for the semantic segmentation of ALS datasets.

# 3  Method

We first describe the design of 2D convolutions and how 2D and 3D convolutions work together in Section 3.1. Then we introduce the SegECC operation that encodes the contextual information at the object level in Section 3.2. Section 3.3 explains how the spatial-channel attention is adjusted to 3D point clouds. Finally, the architecture of LGENet is presented in Section 3.4.

## 3.1  Hybrid convolution block

KPConv (Thomas et al., 2019) is a 3D convolutional kernel whose domain is a spherical 3D space. It has a deformable version that adapts to local geometry in order to enhance the representation of features. However, Thomas et al. (2019) suggest that rigid convolutions perform better than deformable ones on scenes that lack of diversity. As a majority of objects in ALS datasets are buildings, ground and vegetation, pedestrians and road furniture are less likely to be observed, we use rigid convolutions in our experiments. If it is not specified, KPConv represents rigid KPConv in the following paper. In order to extract more representative features for ALS point clouds, a 2D variant of KPConv is applied and is incorporated with 3D KPConv forming a hybrid block.

As ALS point clouds are acquired by airborne LiDAR equipment from a top view and most semantic objects in urban scenes are horizontally distributed on the ground, point cloud variance in the vertical direction (z coordinates) is much less than that in the horizontal plane. Due to this characteristic, 2D convolutions are applied to learn more representative features for urban objects from the point distribution on the horizontal plane. Their effectiveness has been proven by various previous works (Wen et al., 2020; Yang et al., 2017; Zhao et al., 2018), in which point clouds are projected to the horizontal plane and 2D CNN is applied to extract features and predict pointwise semantic labels. In the following section, the mechanism of KPConv is reviewed according to Thomas et al. (2019). Thereafter, how the 2D KPConv works and how the 2D and 3D KPConv are combined are explained.

Given a point $p_i$ from a point cloud $\mathcal{P} \in \mathbb{R}^{N \times 3}$, corresponding features are defined as $f_i$ and $\mathcal{F} \in \mathbb{R}^{N \times C}$. The point convolutional operation at point $p \in \mathbb{R}^{N \times 3}$ is defined as follows:

$$(\mathcal{F} * g)(p) = \sum_{p_i \in N_p} g(p_i - p) f_i \tag{1}$$

Where $g$ is a kernel function and $N_p$ is a set of neighbours around $p$ within a fixed radius $r \in \mathbb{R}$, $N_p = \{p_i \in \mathcal{P} \mid \|p_i - p\| \leq r\}$. The domain of $g$ for 3D KPConv is defined as a 3D ball space centred on $p$, mathematically represented as $D_r^3 = \{x \in \mathbb{R}^{N \times 3} \mid \|x\| \leq r\}$ where $x_i = p_i - p$. Similar to image convolutional kernels, the kernel function $g$ have different weights to different parts inside the kernel domain. Different areas in $D_r^3$ are represented as kernel points $\{\tilde{p}_k \mid k < K\} \subset D_r^3$ and their corresponding weight matrices are denoted as $\{W_{pk} \mid k < K\} \subset \mathbb{R}^{C_{in} \times C_{out}}$, mapping features from dimensions $C_{in}$ to $C_{out}$. The kernel function $g$ for any point $x_i \in D_r^3$ is defined as:

$$g(x_i) = \sum_{k < K} h(x_i, \tilde{p}_k) W_{pk} \tag{2}$$

Here, $h$ is a linear correlation between $\tilde{p}_k$ and $x_i$ whose values are higher when $x_i$ is close to $\tilde{p}_k$.

The 2D kernel function $l$ is quite similar to the 3D one $g$, except $N_q$ and kernel point positions $\tilde{q}_k$ are defined differently.

$$(\mathcal{F} * l)(q) = \sum_{q_i \in N_q} l(q_i - q) f_i \tag{3}$$

$q \in \mathbb{R}^{N \times 2}$ is the projection of $p$ on the XY plane. Instead of searching neighbours among projected 2D points, $N_q$ is the 2D projection of $N_p$. The domain of $l$ is a 2D circular surface, denoted as $D_r^2 = \{y \in \mathbb{R}^{N \times 2} \mid \|y\| \leq r\}$, where $y_i = q_i - q$, the relative position of $q_i$ to the central point $q$. The kernel points in $D_r^2$ are written as $\{\tilde{q}_k \mid k < K\} \subset D_r^2$ and their weight matrices are $\{W_{qk} \mid k < K\} \subset \mathbb{R}^{C_{in} \times C_{out}}$. The 2D kernel function is defined as:

$$l(y_i) = \sum_{k < K} h(y_i, \tilde{q}_k) W_{qk} \tag{4}$$

The distribution of 3D and 2D kernel points are shown in Figure 1. It can be seen that in the 3D kernel, top and bottom kernel points are far away from the object surface so that those points have little to no contribution in describing the local geometry. However, when the distance between a point and a kernel point is only assessed in 2D, most kernel points will have some nearby points on object surfaces. In this way, more kernel points contribute to the feature extraction, leading to more representative features on object surfaces.

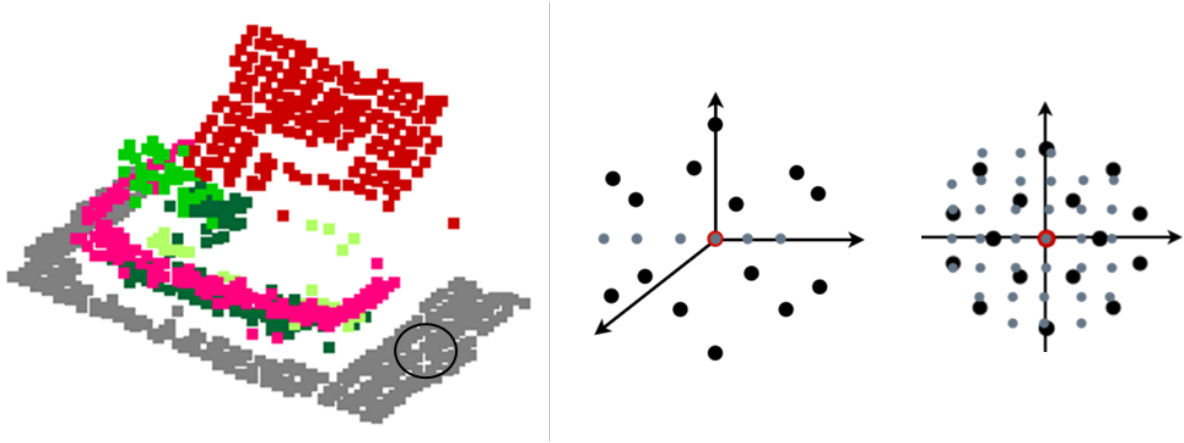

*Figure 1 Illustration of kernel points distribution for 2D and 3D KPConv. Left: an example of ISPRS benchmark dataset. Middle: the front view of fence points and 3D kernel points. Right: the top view of the projected fence points and 2D kernel points. Black dots are kernel points.*

The hybrid KPConv block used in the network is shown in Figure 2. It inherits the ResNet connections from the original block (Thomas et al., 2019). The single 3D-KPConv in the original block is replaced with a 3D-KPConv and a 2D-KPConv. Outputs of those two convolutions are concatenated and passed to as $1 \times 1$ convolution. The effectiveness of the hybrid block is shown in Section 4.1.8.1.

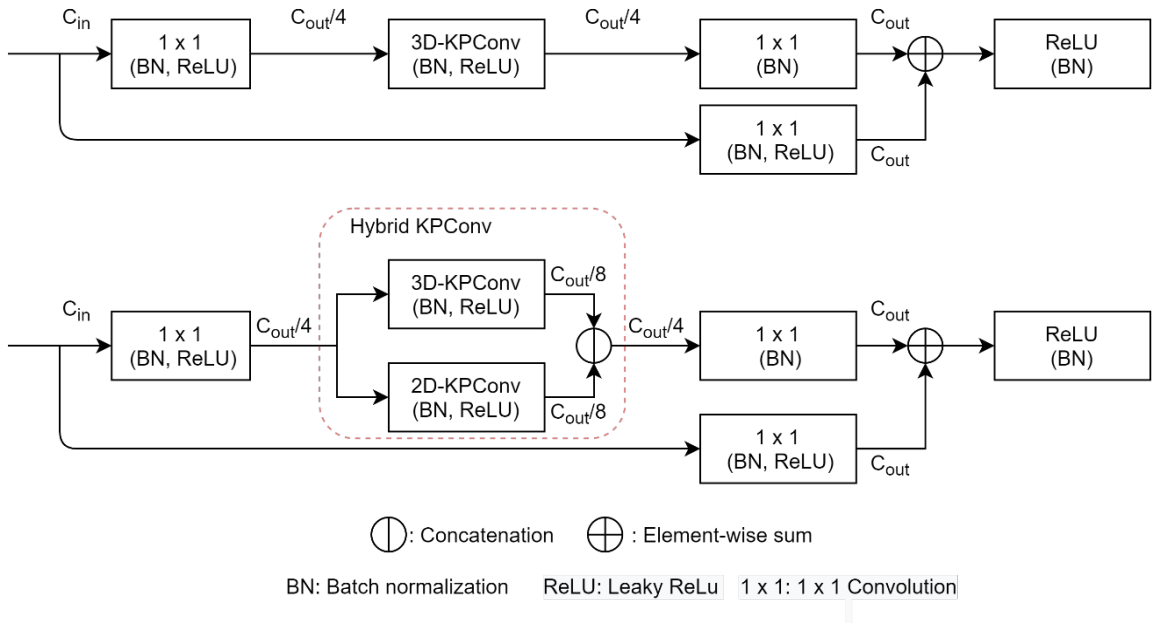

*Figure 2 The convolutional block used in* Thomas et al. (2019) *(top) and the hybrid 2D-3D block used in this paper (bottom). The hybrid block inherits the ResNet connections from the original one. Instead of simply passing through a 3D-KPconv, features are fed to both 2D and 3D KPConvs and outputs of two convolutions are concatenated for the 1×1 convolution.*

## 3.2 SegECC

The pointwise features obtained by hybrid KPConv layers are only representative for local geometry because each convolutional layer only has a local receptive field and pointwise features cannot encode information outside the local region as well as relationships between objects. This is insufficient to explore the inherent structures of large objects and the interactions between objects. Lack of this global context limits the network performance on pointwise prediction for outdoor scenes in ALS point clouds. To achieve better performance, spatial dependencies at the object level from a global perspective should be exploited and integrated with local geometrical features. Inspired by SPG (Landrieu and Simonovsky, 2018), we construct graphs on segments that consist of geometrical homogenous points to capture the relationships among objects. By combining segment features and pointwise features, the network adaptively encodes local-global features, thus achieving better semantic predictions on ALS datasets. The following paragraphs explain how global context is explored at the segment level and how it is aggregated with local features.

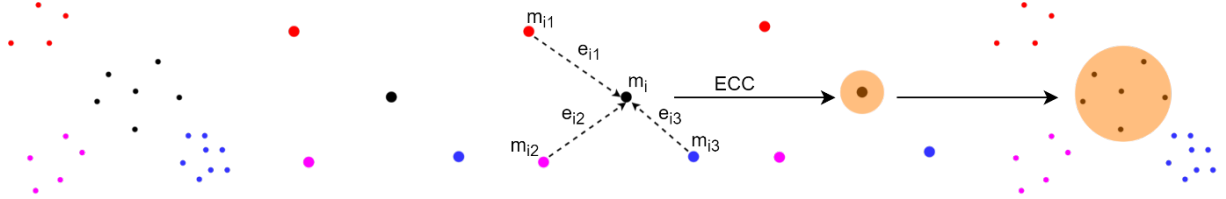

*Figure 3 Steps in SegECC to obtain segment embeddings using edged conditioned convolutions (ECC). Different colors represent different segment labels.*

Figure 3 illustrates the process of the segment-based Edge Conditioned Convolution (SegECC) to obtain global embeddings. Firstly, point clouds are partitioned into segments by an unsupervised algorithm, $L_0$-cut pursuit proposed by Landrieu and Obozinski (2017). The segmentation is performed before the training and is based on predefined geometrical features and intensity. Unlike DNN (Simonovsky and Komodakis, 2017) which dynamically clusters points according to updated features during the training, we use a fixed graph structure in our network and all segment labels are inherited from the initial segmentation. This fixed structure is more computationally efficient because it does not search KNN neighbours in high dimensional feature space for every training iteration. Experimental results in section 4.1.8.2 show the effectiveness of the fixed segment labels. Next, pointwise features obtained from hybrid KPConv are aggregated to node features according to segment labels. Within a segment, the averages of features over all segment points are calculated. For each node, a graph is constructed with all other nodes in the scene. The features at the central node are represented by $m_i$ and its neighbors' features are represented by $m_{ij}$. Edges features are represented by $e_{ij} = m_i - m_{ij}$. Then Edge-Conditioned Convolution (ECC) (Simonovsky and Komodakis, 2017) is used to capture the contextual information among different segments. It can dynamically generate filtering weights are according to $e_{ij}$ and deal with a flexible number of neighbors. The calculation of ECC is shown as the following:

$$m'_i = \frac{1}{|N(i)|} \sum_{j \in N(i)} \Theta(e_{ij}, W_e) m_{ij} \tag{5}$$

$W_e$ are learnable parameters in the multi-layer perceptron $\Theta$. Edge features $e_{ij}$ are processed by $\Theta$ to produce a weight matrix and a matrix-vector multiplication is performed between the weight matrix and neighbouring node features $m_{ij}$. Adaptively weighted $m_{ij}$ are aggregated by the calculation of the mean. Finally, updated node features $m'_i$ are directly back-propagated to each point and concatenated with pointwise features generated from KPconv. Figure 4 demonstrates how SegECC operation is inserted to a hybrid convolution block. The input is the feature obtained from the hybrid KPConv and the output is concatenated with the input for following operations.

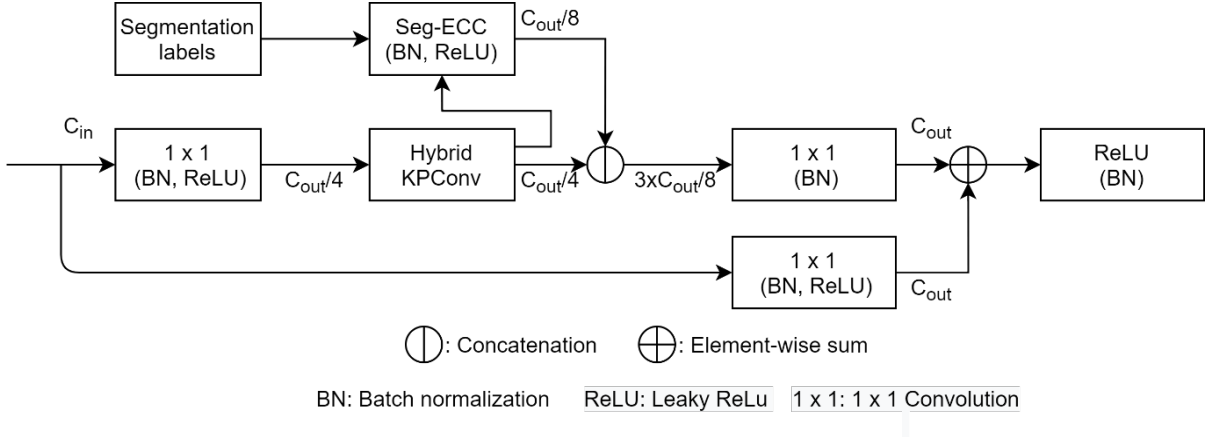

*Figure 4 The structure of the hybrid-SegECC block.*

## 3.3 Spatial-channel attention

Hybrid KPConv and SegECC are proposed to extract representative features at point and object levels. However, it is also necessary to consider global information when determining the semantic label for each point. In semantic segmentation, two points can be the same category even if they are spatially far away. Considering the correlation between these two points in feature space can mutually improve the prediction accuracy. Also, for high dimensional features dependencies between channels exist, which enhance the feature discriminability for different semantic classes. Following the dual attention proposed by Fu et al. (2018) for image semantic segmentation, the spatial-channel attention is proposed for semantic segmentation of ALS point clouds.

In order to model the relationship between any two members in a point cloud, the Spatial-Attention Module is applied to adaptively aggregate pointwise features according to their correlations. As demonstrated in Figure 5, given the feature $F \in \mathbb{R}^{N \times C}$ for a point set, $F$ is fed into two fully connected (FC) layers to obtain transformed features $U$ and $V$, respectively, where $U, V \in \mathbb{R}^{N \times C}$. The spatial attention matrix $SA \in \mathbb{R}^{N \times N}$ is calculated based on the matrix multiplication between the $V$ and the transpose of $U$. Normalized spatial attention map $sa_{ij}$ estimates the impact of point $j$ on point $i$. Similar features of two points give rise to a high correlation between them, contributing to a large value in $sa_{ij}$.

$$sa_{ij} = softmax_j(U_i \cdot V_j) \qquad (6)$$

$F$ is also transformed into a new representation $T \in \mathbb{R}^{N \times C}$ by an FC layer. The matrix multiplication between $SA$ and $T$ is calculated and then times a learnable scale parameter $\alpha$. The output is element-wisely summed with $F$ to obtain the final feature $F_{sa}$.

$$F_{sa} = \alpha \, SA \cdot T + F \qquad (7)$$

The resulting feature $F_{sa}$ encodes the features across all positions and this global view helps similar semantic features to achieve mutual gains, therefore improving the semantic consistency.

Every channel in high level features can be taken as a class-specific response and responses of different semantics are related to each other. Therefore, feature discriminability can be improved by modelling the interdependencies between different channels. Channel attention is employed to exploit channel-wise interdependencies.

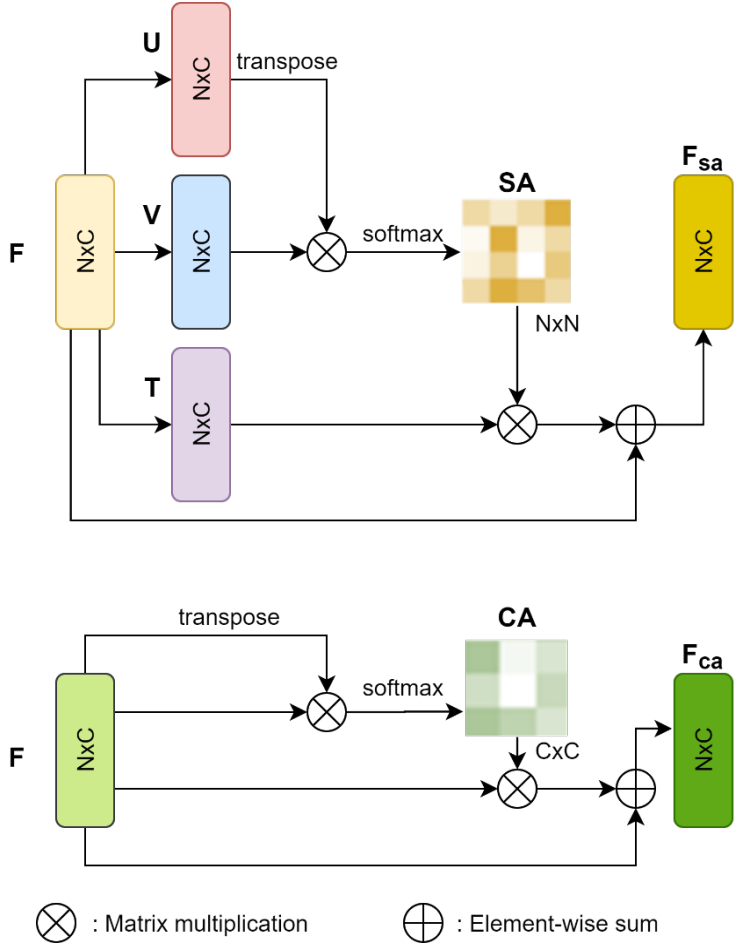

*Figure 5 Structures of spatial attention (top) and channel attention (bottom).*

The structure of the channel attention model is shown in Figure 5. The attention $CA \in \mathbb{R}^{C \times C}$ is directly computed from the feature map $F$ without any transformation. $ca_{ij}$ measures the influence of $j^{th}$ channel on $i^{th}$ channel and $CA$ models the dependency between all channels. Similar to the calculation of the spatial attention module, the resulting feature $F_{ca}$ is obtained by a matrix multiplication between $CA$ and $F$, followed by a scale multiplication and an element-wise summation. The parameter $\beta$ is learned from the training.

$$ca_{ij} = softmax_j(F_i \cdot F_j) \tag{8}$$

$$F_{ca} = \beta \, CA \cdot F + F \tag{9}$$

In order to make full use of the global context, $F_{sa}$ and $F_{ca}$ are summed up and the sum is passed to an FC layer to obtain pointwise semantic labels. With this spatial-channel attention module, pointwise features are updated from a global perspective. The complicated interactions between points are comprehensively learned, contributing to more accurate predictions.

### 3.4 Overall Network architecture

With three blocks introduced above, LGENet that encodes both local and global information can be constructed for semantic segmentation of ALS point clouds. Following the fully convolutional network proposed by Thomas et al. (2019), our network is composed of an encoder and a decoder. As illustrated in Figure 6, the encoder has 5 convolutional layers and each layer consists of two convolutional blocks. We use hybrid KPConv for all convolutional blocks. However, SegECC is only inserted at the second block of the third and fourth layer. According to Feng et al. (2020), using blocks that encode global features in all layers fails to improve model performance because those blocks intensively increase the number of network parameters and it is difficult to achieve a global optima. This is also demonstrated by our experiments shown in Table 5. In order to capture the local geometry at multiple scales, downsampling is used to enlarge the receptive field of convolutions step by step.

In the decoder, nearest upsampling is employed to obtain final pointwise features. Four skip connections are applied to pass intermediate features from encoder to decoder. Those features are concatenated with the upsampled features and then passed to a unary block which is a $1 \times 1$ convolution. At end the of the network, a spatial-channel attention block is stacked to consider global context, thus improving final pointwise semantic predictions.

For the overall network training, we use the weighted cross entropy loss,

$$L = \sum_{i=1}^{N} \sum_{c=1}^{C} w_c(\hat{y}_{ic}\, log\rho_{ic} + (1 - \hat{y}_{ic})log\,(1 - \rho_{ic})) \tag{10}$$

where $\hat{y}_{ic}$ represents whether the ground truth label for $i$ th is in $c$ th category and $\rho_{ic}$ represents the corresponding predicted probability. The weight for each class $w_c$ is calculated by its proportion out of the total number of points,

$$w_c = \frac{1/\gamma_c}{\sum_{c=1}^{C}\frac{1}{\gamma_c}} \quad , \quad \gamma_c = \frac{N_c}{\sum_{c=1}^{C} N_c} \tag{11}$$

where $N_c$ denotes the number of points for $c$th class.

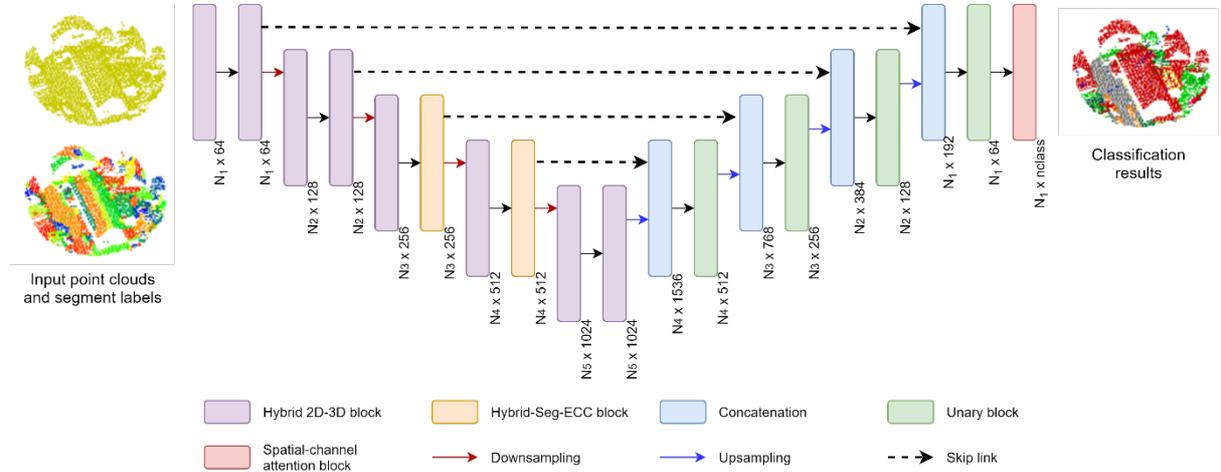

*Figure 6 Illustration of the proposed LGENet architecture for semantic segmentation of ALS point clouds. The encoder consists of hybrid 2D-3D blocks and hybrid-SegECC blocks. The decoder is composed of unary blocks (1×1 convolutions). N1 > N2 > N3 > N4 > N5 denote point numbers. Intermediate features are passed from encoder to decoder through four skip links. The spatial-channel attention block is stacked at the end of the network.*

# 4 Experiments

In this section, experiments are shown to evaluate the effectiveness of the proposed network in two ALS datasets. We compare the performance of our model against that of other the state-of-the-art models on the ISPRS benchmark (Niemeyer et al., 2014). We also conduct a comprehensive ablation study on the ISPRS benchmark (Niemeyer et al., 2014) to show the effectiveness of our proposed method and evaluate how hyperparameters and network structure influence the model performance. Next the DFC2019 dataset (Bosch et al., 2018) is used to further demonstrate the advantages of our method.

## 4.1 Experiments on ISPRS benchmark dataset

### 4.1.1 Dataset

The performance of our method is evaluated by the ISPRS benchmark dataset of 3D labelling (Niemeyer et al., 2014). An overview of the dataset is shown in Figure 7. The benchmark dataset was obtained in August 2008, at Vaihingen, Germany, through a Leica ALS50 system whose mean flying height is 500m and field of view is $45^\circ$. Point clouds were captured with a density of 4 points/m$^2$. Each point has five attributes, namely, XYZ coordinates, intensity values and number of returns. The dataset is labelled into 9 classes, including powerline, low vegetation, impervious surface, car, fence/hedge, roof, façade, shrub, and tree. In ISPRS 3D labelling contest, the point cloud is divided into a training area and a testing area. The training area consists of 753,876 points, dominated by residential buildings. The testing area contains 411,722 points located in the city centre.

Section A                                    Section B

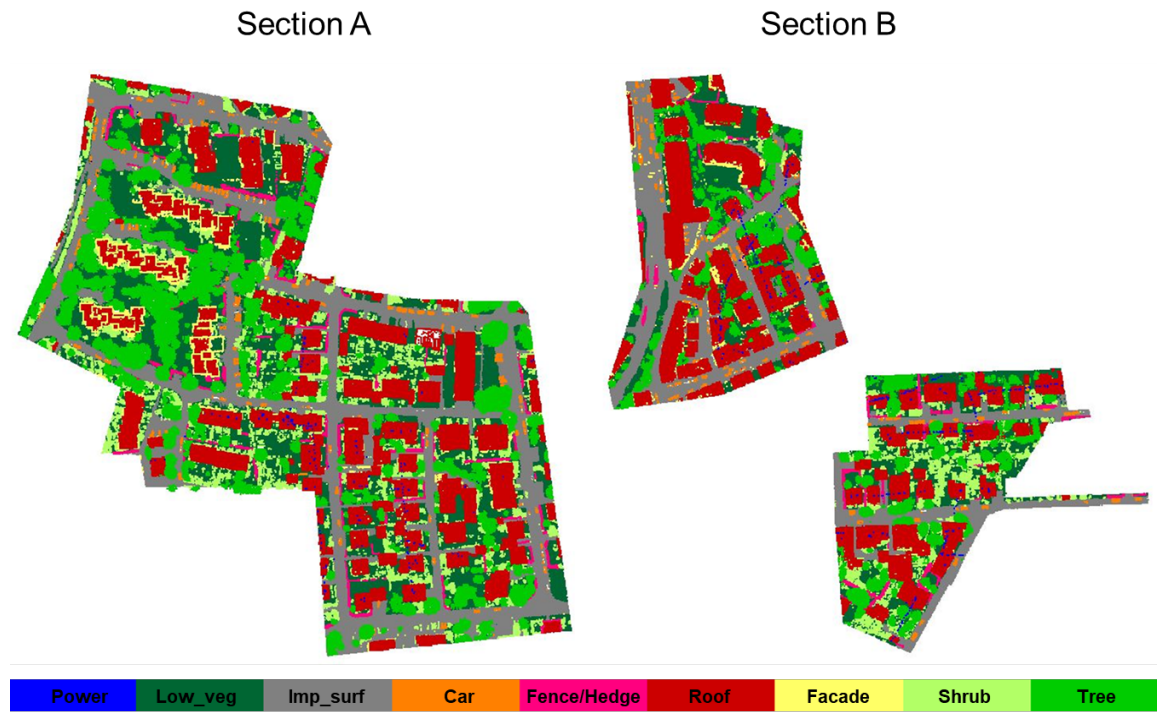

| Power | Low_veg | Imp_surf | Car | Fence/Hedge | Roof | Facade | Shrub | Tree |

*Figure 7 An overview of the ISPRS benchmark dataset. The Section A is used for model training and Section B is used for model evaluation.*

### 4.1.2 Accuracy assessment

Following the evaluation metrics of ISPRS benchmark dataset, we use the average F1 score (Avg. F1) and overall accuracy (OA) to evaluate our method. The OA measures the percentage of correctly predicted points in the total number of test points. F1 score is a statistical metric calculated from precision and recall.

$$precision = TP \ / \ (TP + FP) \qquad (10)$$

$$recall = TP \ / \ (TP + FN) \qquad (11)$$

$$F1 \ score = 2 \times (precision \ \times \ recall) / \ (precision \ + \ recall) \qquad (12)$$

where TP, FN and FP are true positives, false negatives and false positives receptively in a confusion matrix.

### 4.1.3 Preprocessing

The ISPRS benchmark dataset is first segmented by the algorithm proposed by Landrieu and Obozinski (2017) to obtain segment labels required in the SegECC block. We use both XYZ coordinates and intensity as the input of the segmentation algorithm. The most important factor of the segmentation is the regulation strength, determining the coarseness of the final partition. The regulation strength is set to 0.03 in our paper.

When preparing the data for training, we subsample the point clouds with a grid sampling size of 0.24m, in order to deal with the large variation in point density in ALS point clouds. Spheres are randomly selected from the subsampled point clouds and fed into the network. The radius of the input sphere is taken as 24m. We use intensity, absolute Z- coordinates and normalized Z- coordinates within the sphere as input features. For data augmentation, the input sphere is randomly rotated around the Z-axis to improve the network robustness to orientation. Also, random noises are added to XYZ coordinates with a σ of 4 cm which is chosen empirically and will not significantly modify the geometry of target objects.

### 4.1.4 Network implementation

As mentioned in section 4.1.3, input point clouds are downsampled in different layers. Table 1 shows the grid size of the downsampling and the size of convolution kernels from layer 1 to layer 5. The convolution radius is 2.5 times the grid size in the corresponding layer. For example, the input of the first convolutional layer is subsampled with the grid size of 0.24m and the radius of the convolution in the first layer is 0.6m The number of kernel points in 3D KPConv is 15 and that of 2D KPConv is 17. Kernel points are initialized by the energy function proposed by Thomas et al. (2019) to ensure they are far from each other inside a given sphere (or circle for the 2D kernel).

*Table 1 Subsampling grid size and convolution radius in different layers.*

| Layer | 1 | 2 | 3 | 4 | 5 |
|---|---|---|---|---|---|
| Subsampling grid size (m) | 0.24 | 0.48 | 0.96 | 1.92 | 3.84 |
| Convolution radius (m) | 0.6 | 1.2 | 2.4 | 4.8 | 9.6 |

### 4.1.5 Training and testing

The proposed network is implemented based on the PyTorch framework (Paszke et al., 2019), trained with a Geforce RTX 2080 Ti GPU. Stochastic gradient descent (SGD) optimiser is applied to optimize network weights. The weighted cross entropy loss function is applied to rebalance imbalanced data. During the training, we take 2000 iterations as one epoch. The learning rate starts from 0.001 with a decay rate of 0.9 at every 5 epochs. The model is trained for 60 epochs until the convergency is achieved. For testing, we randomly select spheres in the test area and each point is repeatedly fed into the network at least 20 times to obtain averages predictive probability. This repetition is to avoid the misclassification on points near the sphere boundary whose geometry may be incomplete.

### 4.1.6 Classification results

Qualitative results are shown in Figure 8 and the corresponding error map is demonstrated in Figure 9. It can be seen that the proposed LGENet correctly predicts most of the points in the testing area (Figure 9). As shown in Figure 8, car and façade points are well predicted, even though they have less instances in the whole dataset. Also, the LGENet can effectively identify powerline points although they are sparsely distributed above all other classes.

Classification results are quantitively present by the confusion matrix in Table 2. LGENet can effectively recognize most of the classes with an overall accuracy of 0.845. Powerline, low vegetation, impervious surface, roof and tree are well recognized. The worst classification result lies in the fence/hedge, most likely it will be predicted as shrub points according to the confusion matrix. This confusion is due to these classes being similar in geometry, pulse intensity and spatial distribution. In addition, façade points are also likely to be identified as shrub points since points are relatively sparse on façade in ALS dataset and they are difficult to separate if shrub points are very close to the building.

*Table 2 Confusion matrix of our proposed network on ISPRS benchmark dataset. Precision, recall and F1 score are reported for each class. The OA is 0.845 and the average F1 score is 0.737.*

|  | Power | Low_veg | Imp_surf | Car | Fence/Hedge | Roof | Facade | Shrub | Tree |
|---|---|---|---|---|---|---|---|---|---|
| Power | 459 | 2 | 0 | 0 | 0 | 95 | 16 | 1 | 27 |
| Low_veg | 0 | 83454 | 5870 | 61 | 263 | 1201 | 391 | 5008 | 2442 |
| Imp_surf | 0 | 9318 | 91972 | 44 | 18 | 301 | 35 | 296 | 2 |
| Car | 0 | 206 | 144 | 2612 | 84 | 112 | 7 | 518 | 25 |
| Fence/Hedge | 0 | 871 | 103 | 5 | 2063 | 188 | 33 | 3217 | 942 |
| Roof | 112 | 3883 | 114 | 3 | 60 | 101146 | 1486 | 1229 | 1015 |
| Facade | 14 | 776 | 77 | 34 | 34 | 1243 | 6583 | 1863 | 600 |
| Shrub | 1 | 4682 | 75 | 57 | 97 | 1281 | 368 | 14319 | 3938 |
| Tree | 14 | 1391 | 15 | 4 | 126 | 938 | 200 | 6144 | 45394 |
|  |  |  |  |  |  |  |  |  |  |
| Precison | 0.765 | 0.798 | 0.935 | 0.926 | 0.752 | 0.950 | 0.722 | 0.439 | 0.835 |
| Recall | 0.765 | 0.846 | 0.902 | 0.704 | 0.278 | 0.928 | 0.587 | 0.577 | 0.837 |
| F1 | 0.765 | 0.821 | 0.918 | 0.800 | 0.406 | 0.938 | 0.647 | 0.499 | 0.836 |

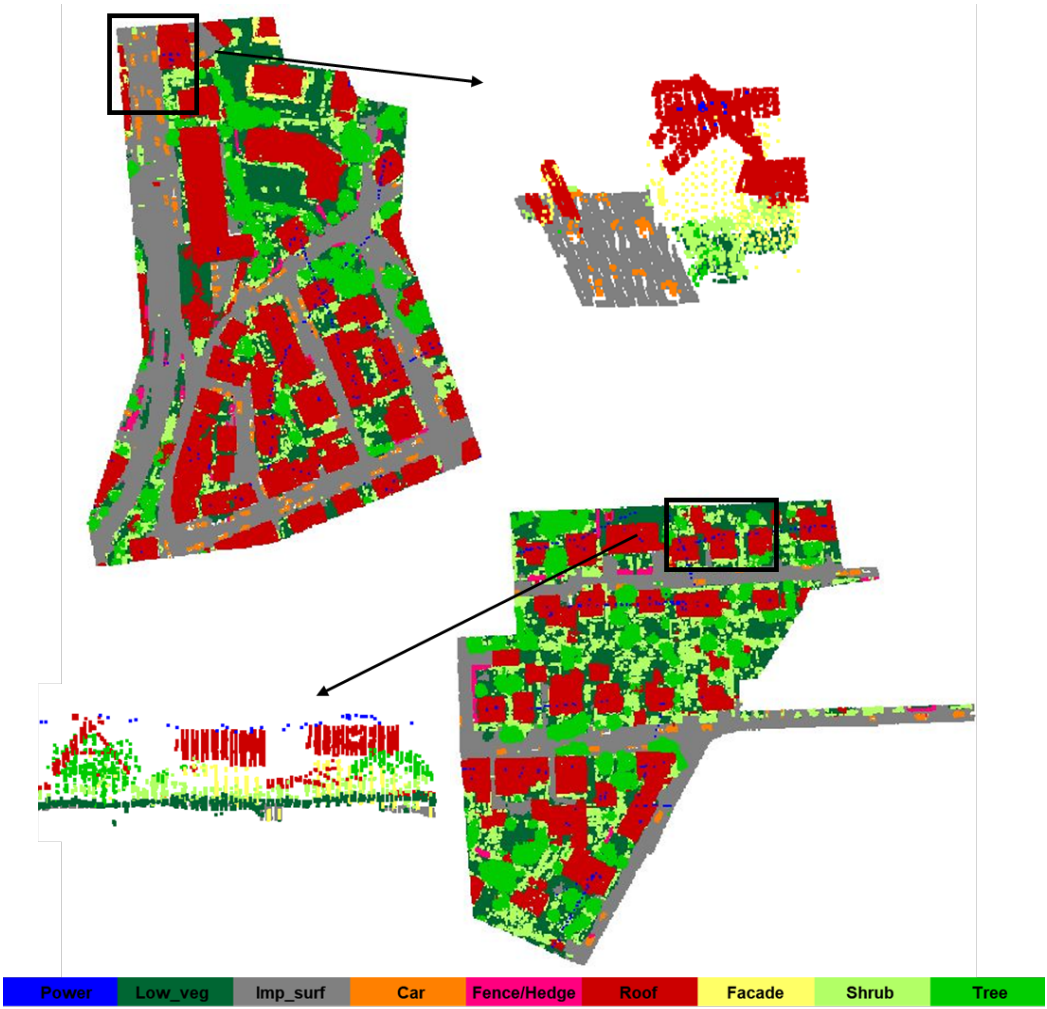

| Power | Low_veg | Imp_surf | Car | Fence/Hedge | Roof | Facade | Shrub | Tree |

*Figure 8 Classification results of our LGENet on the ISPRS benchmark dataset.*

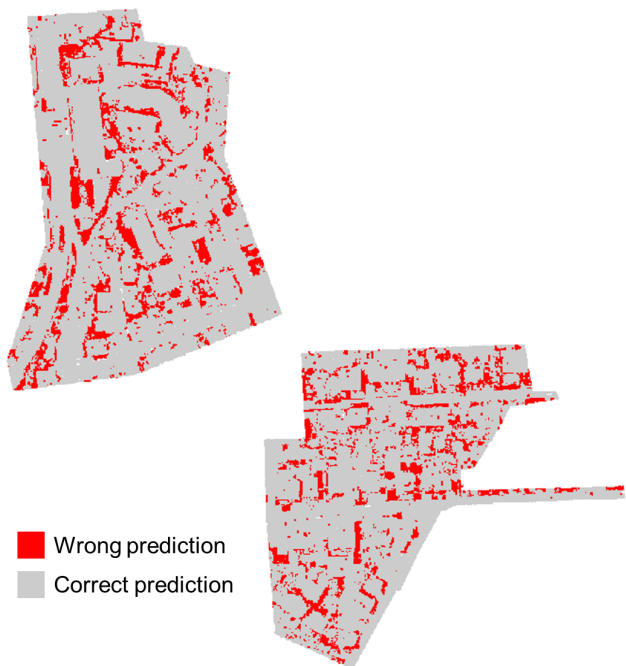

■ Wrong prediction
■ Correct prediction

*Figure 9 The error map of our LGENet on the ISPRS benchmark dataset.*

### 4.1.7 Comparison to state-of-the-art methods

We quantitatively compare our LGENet to other state-of-the-art models on the ISPRS benchmark dataset in Table 3. LUH (Niemeyer et al., 2016) relies on handcrafted features and applies a two-layer hierarchical CRF at point and segment level. Other methods are based on deep learning, namely, WhuY4 (Yang et al., 2018), RIT_1 (Yousefhussien et al., 2018), alsNet (Winiwarter et al., 2019), A-XCRF (Arief et al., 2019), D-FCN (Wen et al., 2020), and Li et al. (2020).

Comparing with all the above methods, LGENet achieves superior classification performances in terms of the average F1 score (0.737). Nevertheless, the OA of LGENet is slightly lower than the highest OA (0.850) obtained by A-XCRF. One explanation for this is that we apply a weighted loss to balance the imbalanced class distribution in the ISPRS benchmark dataset. The focusing on OA leads to bias on dominant categories and ignores minority classes in the dataset. Therefore, it is more meaningful to evaluate model performance by average F1 scores that equally reflect the model performance for all categories. Our LGENet significantly improves the baseline (KPConv) by 0.031 in the average F1 score and OA in 0.028. As the data processing and hyper-parameter settings are the same for LGENet and the baseline network, the accuracy improvement is a result of our network design.

*Table 3 Quantitative comparisons between our LGENet and other models on the ISPRS benchmark dataset. The F1 scores for different classes are shown in the first nine columns and the OA and the average F1 score are shown in the last two columns. The boldface text means the highest value in the column.*

|  | Power | Low_veg | Imp_surf | Car | Fence/Hedge | Roof | Facade | Shrub | Tree | Avg. F1 | OA |
|---|---|---|---|---|---|---|---|---|---|---|---|
| LUH | 0.596 | 0.775 | 0.911 | 0.731 | 0.340 | 0.942 | 0.563 | 0.466 | 0.831 | 0.684 | 0.816 |
| WhuY4 | 0.425 | **0.827** | 0.914 | 0.747 | **0.537** | 0.943 | 0.531 | 0.479 | 0.828 | 0.692 | 0.849 |
| RIT_1 | 0.375 | 0.779 | 0.915 | 0.734 | 0.180 | 0.940 | 0.493 | 0.459 | 0.825 | 0.633 | 0.816 |
| alsNet | 0.701 | 0.805 | 0.902 | 0.457 | 0.076 | 0.931 | 0.473 | 0.347 | 0.745 | 0.604 | 0.806 |
| A-XCRF | 0.630 | 0.826 | **0.919** | 0.749 | 0.399 | **0.945** | 0.593 | 0.507 | 0.827 | 0.711 | **0.850** |
| D-FCN | 0.704 | 0.802 | 0.914 | 0.781 | 0.370 | 0.930 | 0.605 | 0.460 | 0.794 | 0.707 | 0.822 |
| Li et al. (2020) | 0.754 | 0.820 | 0.916 | 0.778 | 0.442 | 0.944 | 0.615 | 0.496 | 0.826 | 0.732 | 0.845 |
| KPConv | 0.735 | 0.787 | 0.880 | 0.794 | 0.330 | 0.942 | 0.613 | 0.457 | 0.820 | 0.706 | 0.817 |
| Ours | **0.765** | 0.821 | 0.918 | **0.800** | 0.406 | 0.938 | **0.647** | **0.499** | **0.836** | **0.737** | 0.845 |

### 4.1.8 Ablation study

#### 4.1.8.1 Effectiveness of hybrid convolution

To justify the importance of 2D KPConv in semantic segmentation of ALS point clouds, we conduct experiments to compare and contrast models with and without 2D convolutions. We also evaluate how the model performance changes with a different number of kernel points in 2D convolutions. Furthermore, we test the model performance when only using 2D convolutions and when searching neighbours among projected 2D points in the 2D KPConv of hybrid blocks.

Table 4 presents the quantitative results using different convolutions. The original 3D KPConv is used as the baseline. Although deformable KPConv kernels (Thomas et al., 2019) are adaptive to object surface and enhance descriptive power of output features, they fail in the experiments on the ISPRS benchmark dataset (last row in Table 4). According to Thomas et al. (2019), this is because the dataset has only 9 classes and lacks object diversity comparing to other datasets with more complex scenes. By comparing the baseline to the only-2D network, it can be seen that although the 2D convolutions lead to low F1 scores in most of the semantic classes, the only-2D network outperforms the baseline with 3D convolutions for the fence/hedge class which is a very difficult class for other methods. A possible explanation for this could be that fence/hedge are elongated structures distributed on XY plane and fixed kernel points on XY planes contributes to better representations. For rigid 3D convolutions, kernel points are distributed in the sphere and very limited kernel points are located near the XY plane, resulting in the failure on those elongated structures distributed on the ground. When combining 2D and 3D KPConvs, we could see better average F1 and overall accuracy (OA) are obtained with more kernel points in 2D KPConv (Hybrid(5), Hybrid(9) and Hybrid(17) in Table 4. Using 2D KPConv leads to more confusion between powerline points and roof points because projecting points to an XY plane is likely to cause the overlap between the powerline points and roof points and responses of 2D kernels for these two classes can be very similar, while this adverse impact is relieved when using more kernel points to enhance the descriptive power of convolutions. The average F1 and OA achieved are 0.717 and 0.837 respectively when the hybrid convolution layers have 17 kernel points in the 2D convolution. Comparing to the baseline network, this combination significantly improves the F1 scores in fence/hedge and shrub by solving the confusion between them. We also show the results of searching neighbours among projected 2D points in the 2D KPConv of hybrid blocks in the sixth row of Table 4. When comparing to the neighbour searching strategy mentioned in section 3.1 (Hybrid(17)), F1 scores for all categories are lower, especially powerline. This is probably because powerlines always hang over all other objects and searching neighbours among projected 2D points brings irrelevant points like impervious surface and façade points which only give noises and have no contribution to classification results.

*Table 4 Quantitative results (F1 scores) of hybrid KPConv with different numbers of kernel points in the 2D convolution on ISPRS benchmark dataset. Here, we fixed the number of kernel points in 3D convolution as 15. The baseline network uses rigid 3D KPConv proposed by Thomas et al. (2019). The hybrid models involve 2D KPConv in all convolutional layers in the network. The number in the bracket represents the number of kernel points in 2D KPConv.*

| | Power | Low_veg | Imp_surf | Car | Fence/ Hedge | Roof | Facade | Shrub | Tree | Avg. F1 | OA |
|---|---|---|---|---|---|---|---|---|---|---|---|
| Base | 0.735 | 0.787 | 0.880 | 0.794 | 0.330 | 0.942 | 0.613 | 0.457 | 0.820 | 0.706 | 0.817 |
| Hybrid (5) | 0.657 | 0.806 | 0.909 | 0.756 | 0.365 | 0.938 | 0.627 | 0.486 | 0.807 | 0.706 | 0.829 |
| Hybrid (9) | 0.693 | 0.803 | 0.900 | 0.762 | 0.363 | 0.937 | 0.632 | 0.497 | 0.824 | 0.712 | 0.829 |
| Hybrid (17) | 0.703 | 0.811 | 0.908 | 0.757 | 0.381 | 0.939 | 0.632 | 0.495 | 0.826 | **0.717** | **0.837** |
| Only 2D (17) | 0.637 | 0.741 | 0.853 | 0.794 | 0.389 | 0.862 | 0.629 | 0.380 | 0.793 | 0.675 | 0.777 |
| Hybrid (17) 2D neighbours | 0.651 | 0.801 | 0.889 | 0.755 | 0.347 | 0.932 | 0.625 | 0.460 | 0.809 | 0.697 | 0.823 |
| Deformable kernels | 0.604 | 0.743 | 0.879 | 0.734 | 0.403 | 0.941 | 0.595 | 0.453 | 0.820 | 0.686 | 0.812 |

### 4.1.8.2 Effectiveness of SegECC convolution

To take advantage of the SegECC operation, we place the SegECC at different hybrid convolutional layers in the network architecture and quantitative results are shown in Table 5. As shown in the first and second columns in Table 3, adding SegECC at 1 to 4 layers and 2 to 4 layers fails to improve the network performance in terms of average F1 and OA, compared with the network only using hybrid convolutional layers. This is in accordance with observations obtained by Feng et al. (2020) that adding more layers to encode global context in an encoder and decoder network deteriorates model performance. One possible explanation for this drop is that more SegECC blocks raise the number of network parameters and therefore the network fails to achieve global optima. When only inserting the SegECC at the fourth layer, the average F1 score is slightly increased. F1 scores on most of the classes are quite similar to the results of the hybrid network, like low-veg, fence/hedge and roof. The model performance achieves the best by adding the SegECC at layer 3 and layer 4. It outperforms the hybrid network (No SegECC in Table 5) in terms of average F1 score and OA by 0.007 and 0.006. The most significant increase lies in the powerline, façade and fence/hedge which only takes a small proportion of the training data and are difficult to predict. This suggests that the global contextual information captured at an object level is valuable for these classes. This is probably because distributions of these objects are quite unique in urban scenes. Façades exist with roofs. Powerlines are always above all other objects and fences/hedges always surround buildings. However, this global context fails to solve the confusion between shrub and tree because shrub and tree objects are always mixed distributed in urban scenes. Thus, exploiting global context at the object level is limited in distinguishing these two classes.

*Table 5 Quantitative comparison of classification results using SegECC operations at different hybrid convolutional layers on ISPRS benchmark dataset. The last row shows the results of the network only use hybrid convolutional layers without SegECC operation.*

|  | Power | Low_veg | Imp_surf | Car | Fence/Hedge | Roof | Facade | Shrub | Tree | Avg. F1 | OA |
|---|---|---|---|---|---|---|---|---|---|---|---|
| 1,2,3,4 | 0.651 | 0.806 | 0.902 | 0.704 | 0.310 | 0.936 | 0.631 | 0.417 | 0.795 | 0.683 | 0.821 |
| 2,3,4 | 0.725 | 0.815 | 0.911 | 0.750 | 0.377 | 0.925 | 0.615 | 0.432 | 0.797 | 0.705 | 0.829 |
| 3,4 | 0.740 | 0.819 | 0.916 | 0.749 | 0.420 | 0.941 | 0.649 | 0.466 | 0.814 | **0.724** | **0.843** |
| 4 | 0.773 | 0.808 | 0.915 | 0.745 | 0.381 | 0.936 | 0.642 | 0.446 | 0.815 | 0.718 | 0.834 |
| No SegECC | 0.703 | 0.811 | 0.908 | 0.757 | 0.381 | 0.939 | 0.632 | 0.495 | 0.826 | 0.717 | 0.837 |

Table 6 lists results of ISPRS benchmark dataset when SegECC operation uses different segmentation results obtained from different values of the regularization factor in the segmentation algorithm. Qualitative results are demonstrated in Figure 10. It can be seen that with a large regularization factor (0.1), coarse segmentation results are obtained. This under-segmentation cannot separate delicate structures like fence/hedge from other objects and therefore prevents the network from learning interactions among different objects, resulting in poor semantic segmentation results shown in the last row in Table 6. Smaller regularization factors (0.01 and 0.03) produce more detailed segmentation which allows the network to capture the interactions among different segments within a single object and relationships between different objects, thus, improving semantic segmentation results. Comparing the results of two small regularization factors, 0.03 gives better results than 0.01 in terms of OA and it achieves better accuracy in low vegetation, impervious surface, roof and façade. These classes are large objects and segmentation results of 0.01 are too fragmented to assist the network to learn better representations. Therefore, we use 0.03 as the regularization factor to obtain segmentation labels before the network training.

*Table 6 Comparison of model performance on ISPRS benchmark dataset when using different regularization factors in the segmentation algorithm.*

| Reg. factor | Power | Low_veg | Imp_surf | Car | Fence/Hedge | Roof | Facade | Shrub | Tree | Avg. F1 | OA |
|---|---|---|---|---|---|---|---|---|---|---|---|
| 0.01 | 0.717 | 0.811 | 0.910 | 0.778 | 0.397 | 0.937 | 0.647 | 0.476 | 0.818 | 0.721 | 0.835 |
| 0.03 | 0.740 | 0.819 | 0.916 | 0.749 | 0.420 | 0.941 | 0.649 | 0.466 | 0.814 | **0.724** | **0.843** |
| 0.10 | 0.619 | 0.767 | 0.843 | 0.663 | 0.209 | 0.929 | 0.625 | 0.417 | 0.797 | 0.652 | 0.795 |

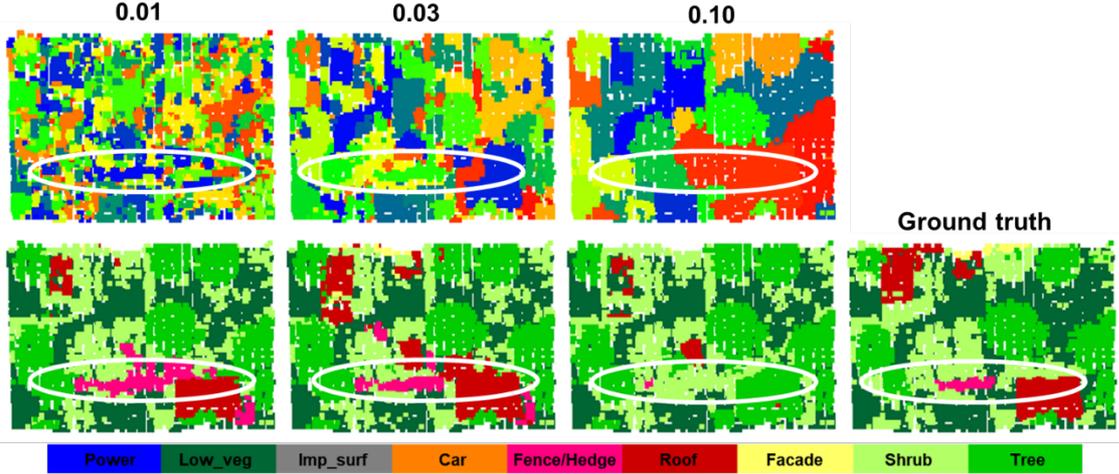

*Figure 10 Qualitative classification results on ISPRS benchmark dataset with different regularization factors. The top row shows the segmentation results obtained from the unsupervised segmentation algorithm and different colours represent different segments. The bottom row presents the correspond semantic segmentation results and ground truth.*

In order to improve the model robustness and save GPU memory, we randomly select edges instead of using all edges during the training. Table 7 presents the classification results using a different number of edges in the SegECC operation. For simplicity, we select the same amount of edges in SegECC regardless of the layer. Considering the GPU memory and neighborhood sizes in layer 3 and layer 4 (Figure 11), we try 3 values for the number of selected edges, namely 40, 80 and 120. It can be seen from Table 7 that the use of too many edges cannot improve model performance due to the overfitting while using only a few edges is insufficient to exploit contextual information. Therefore, we randomly select 80 edges for SegECC in layer 3 and layer 4.

*Table 7 Comparison of model performance on ISPRS benchmark dataset with a different number of edges selected in SegECC operation.*

|  | Power | Low_veg | Imp_surf | Car | Fence/Hedge | Roof | Facade | Shrub | Tree | Avg. F1 | OA |
|---|---|---|---|---|---|---|---|---|---|---|---|
| 40 | 0.707 | 0.809 | 0.905 | 0.727 | 0.402 | 0.939 | 0.643 | 0.448 | 0.814 | 0.711 | 0.831 |
| 80 | 0.740 | 0.819 | 0.916 | 0.749 | 0.420 | 0.941 | 0.649 | 0.466 | 0.814 | **0.724** | **0.843** |
| 120 | 0.696 | 0.801 | 0.903 | 0.760 | 0.359 | 0.930 | 0.625 | 0.437 | 0.801 | 0.701 | 0.822 |

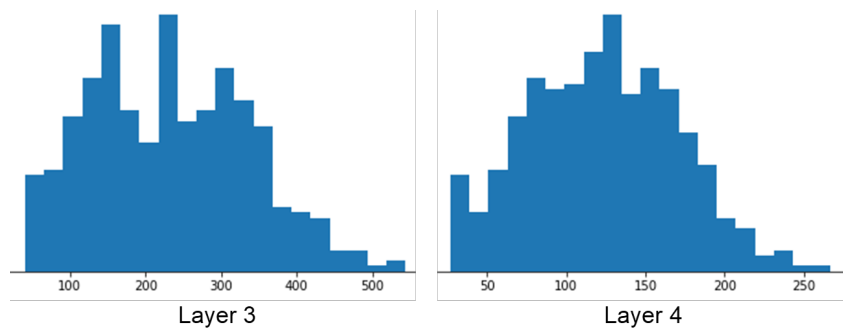

*Figure 11 Distribution of neighborhood size for Vaihingen dataset.*

### 4.1.8.3   Effectiveness spatial-channel attention

Quantitative results of the model using spatial-channel attention on ISPRS benchmark dataset are shown in Table 8. Using spatial-channel attention does not significantly improve OA but increases average F1 score from 0.724 to 0.737. The F1 score in powerline, car and shrub increase by 0.025, 0.051 and 0.033 respectively. Figure 12 shows that with the spatial-channel attention, powerline and car points can be corrected by taking global contextual information. Also, the spatial-channel eliminates "salt and pepper" effects in the classification result. In the bottom row of Figure 12, it removes isolated façade points to make results more consistent with surrounding points, although the model with the spatial-channel attention wrongly predicts hedge points to be shrub and tree points. These incorrect predictions lead to the slight decrease in the F1 score for fence/hedge shown in Table 8.

*Table 8 Comparison of model performance with spatial-channel attention and without spatial-channel attention.*

|  | Power | Low_veg | Imp_surf | Car | Fence/Hedge | Roof | Facade | Shrub | Tree | Avg. F1 | OA |
|---|---|---|---|---|---|---|---|---|---|---|---|
| Without spatial-channel attention | 0.740 | 0.819 | 0.916 | 0.749 | 0.420 | 0.941 | 0.649 | 0.466 | 0.814 | 0.724 | 0.843 |
| With spatial-channel attention | 0.765 | 0.821 | 0.918 | 0.800 | 0.406 | 0.938 | 0.647 | 0.499 | 0.836 | 0.737 | 0.845 |

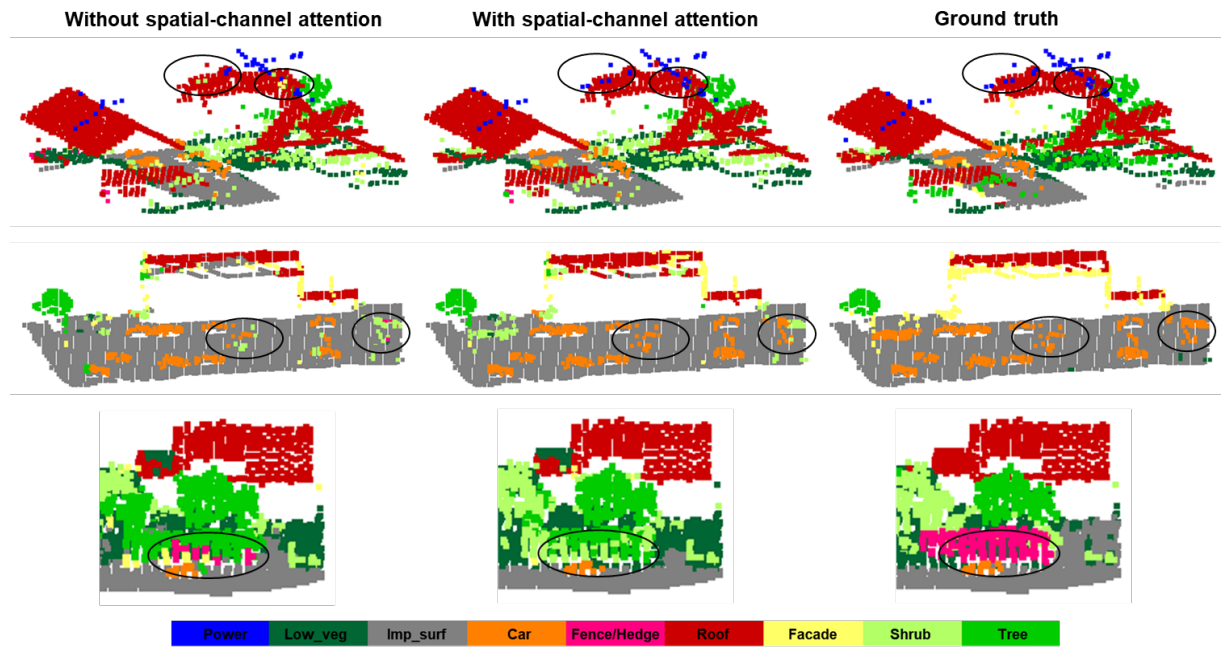

*Figure 12 Qualitative comparison of model performance with spatial-channel attention and without spatial-channel attention.*

## 4.2 Experiments on DFC2019 dataset

### 4.2.1 Dataset

We also evaluate our LGENet by another ALS dataset published by the IEEE Geoscience and Remote Sensing Society (GRSS) for the Data Fusion Contest in 2019 (DFC2019) (Bosch et al., 2018). The DFC2019 dataset covers large-scale urban areas, about 100 km$^2$, in two large cities, namely, Omaha, Nebraska and Jacksonville, Florida in the Unites States. The ALS point clouds are captured with an aggregate nominal pulse spacing of 0.8m and contain about 200 million points in total. For each point, not only XYZ coordinates but also the intensity and return number are available. The point clouds are manually labelled into five classes, namely ground, high vegetation, building, water and bridge deck. We use the same hyperparameters as those for experiments on ISPRS benchmark dataset, except the radius of the input sphere and the grid cell size for subsampling which are set as 48m and 0.48m respectively. This is because DFC point clouds are sparse and only large objects are required to be predicted like building and bridge deck. Furthermore, the model is trained for 150 epochs, as DFC2019 is a large dataset that requires more training to achieve convergence.

### 4.2.2 Classification results

Table 9 and Figure 13 present the classification results of our method on the DFC2019 dataset. The use of hybrid convolutions only give rise to the F1 score increase in building, water and ridge deck by 0.032, 0.009 and 0.041 respectively compared to the baseline KPConv. The average F1 score is therefore increased by 0.017. When adding SegECC to the network (third row in Table 9), it can be seen that the F1 score of the bridge deck rises by 0.056 and that of water slightly improves. The contextual information at the segmentation level helps to correct the building points near the bridge deck and ground points on the river (Figure 13). The effectiveness of spatial-channel attention at the end of the network can be noticed by comparing the third and fourth rows in Table 9. Pointwise predictions are further corrected by considering contextual information from a global perspective. In order to demonstrate the advantages of our LGENet, we also compare our results on DFC2019 with others' results, namely PointNet++ (Qi et al., 2017b) PointSIFT (Jiang et al., 2018) PointCNN (Li et al., 2018) and D-FCN (Wen et al., 2020), shown in Table 10. Our LGENet achieves not only the highest overall accuracy but also the best F1 score for all categories.

*Table 9 Quantitative classification results of different models on the DFC2019 dataset. The first five columns show F1 scores for five classes. The last two columns list average F1 score and OA respectively. The boldface text demonstrates the highest value in the column.*

|  | Ground | High Vegetation | Building | Water | Bridge Deck | Avg. F1 | OA |
|---|---|---|---|---|---|---|---|
| Baseline (KPConv) | 0.991 | 0.975 | 0.893 | 0.434 | 0.694 | 0.797 | 0.978 |
| Hybrid | 0.992 | 0.974 | 0.925 | 0.444 | 0.735 | 0.814 | 0.980 |
| Hybrid-SegECC | 0.993 | 0.979 | 0.924 | 0.447 | 0.791 | 0.827 | 0.983 |
| Hybrid-SegECC-DA | 0.993 | 0.983 | 0.928 | 0.474 | 0.791 | **0.834** | **0.984** |

*Table 10 Quantitative comparisons between other methods and our LGENet on the DFC2019 dataset. The first five columns show F1 scores for five classes. The last two columns list average F1 scores and OA. The boldface text means the highest value in the column.*

|  | Ground | High Vegetation | Building | Water | Bridge Deck | Avg. F1 | OA |
|---|---|---|---|---|---|---|---|
| PointNet++ | 0.983 | 0.958 | 0.797 | 0.044 | 0.073 | 0.571 | 0.927 |
| PointSIFT | 0.986 | 0.970 | 0.855 | 0.464 | 0.604 | 0.776 | 0.940 |
| PointCNN | 0.987 | 0.972 | 0.849 | 0.441 | 0.653 | 0.780 | 0.938 |
| D-FCN | 0.991 | 0.981 | 0.899 | 0.450 | 0.730 | 0.810 | 0.956 |
| Ours | **0.993** | **0.983** | **0.928** | **0.474** | **0.791** | **0.834** | **0.984** |

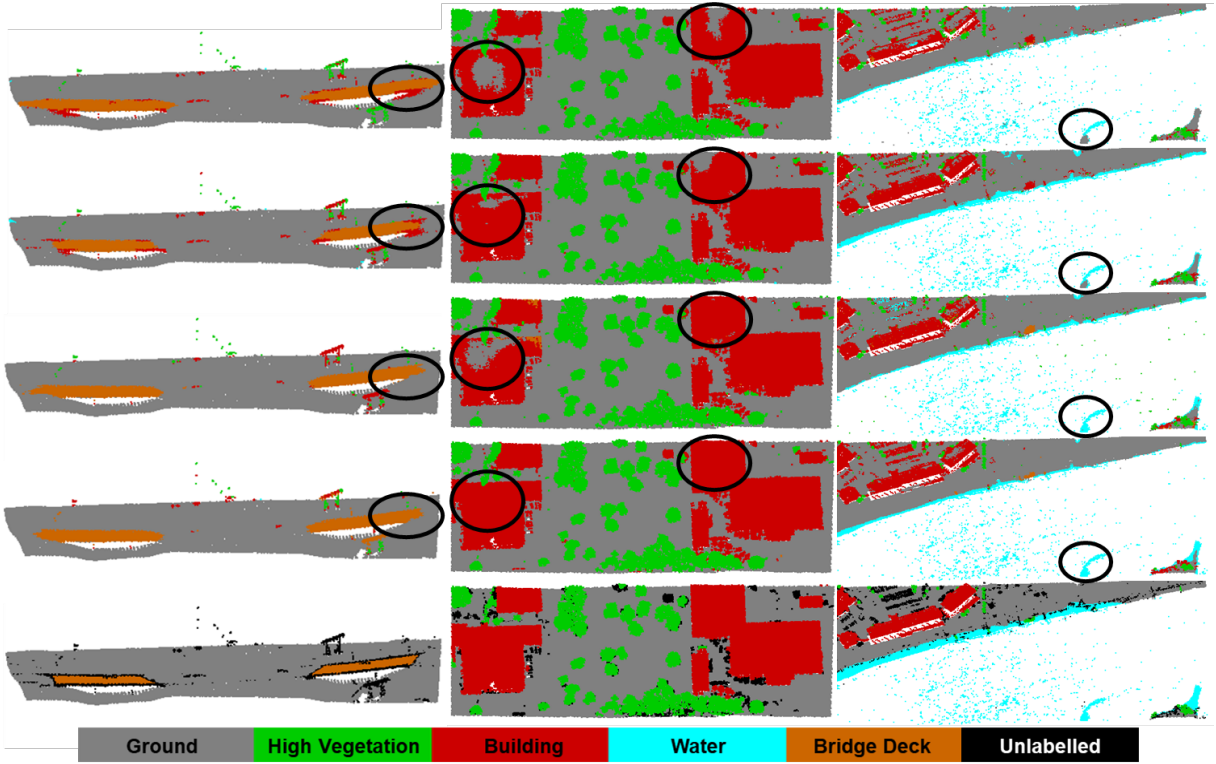

*Figure 13 Some examples of classification results on the DFC2019 dataset obtained from different models. First row: baseline (KPConv), second row: hybrid convolution, third row: hybrid convolution with SegECC operations, fourth row: hybrid convolution with SegECC operations and spatial-channel attention at the end of the network, fifth row: ground truth.*

## 5 Conclusion

We proposed a novel network LGENet for semantic segmentation of ALS datasets. The LGENet learns representative features from local to global and exploits contextual information at both object and point levels. We first add 2D point convolutions to the 3D point convolutions of KPConv forming a hybrid block in order to enhance the learning of the representative local geometry, especially for elongated objects distributed on the horizontal plane. Next, segmented-based edge conditioned convolution (SegECC) is inserted at the end of the hybrid block to encode the context at the object level. Segment labels used in SegECC are obtained before the training from an unsupervised segmentation algorithm, while the segment features are dynamically calculated from changing pointwise features during the training. We finally add a spatial-channel attention module to further improve the semantic predictions by considering global relationships between points and interdependencies between channels.

We verify the advantages of our proposed method by comprehensive ablation experiments on the ISPRS benchmark dataset. LGENet outperforms the baseline model, KPConv (Thomas et al., 2019), by 0.031 in OA and 0.028 in average F1 score. When comparing to other state-of-the-art models, LGENet achieves the best accuracy in terms of average F1 score (0.737). The OA is comparable with the published results obtained from those current leading models. Furthermore, we conduct experiments on the DFC2019 dataset and LGENet produces more accurate pointwise semantic predictions in terms of OA (0.984) and average F1 score (0.834), compared to the baseline and other leading models. This further demonstrates the advantage of our proposed LGENet.

# References

Arief, H.A., Indahl, U.G., Strand, G.H., Tveite, H., 2019. Addressing overfitting on point cloud classification using Atrous XCRF. ISPRS Journal of Photogrammetry and Remote Sensing 155, 90–101. https://doi.org/10.1016/j.isprsjprs.2019.07.002

Armeni, I., Sener, O., Zamir, A.R., Jiang, H., Brilakis, I., Fischer, M., Savarese, S., 2016. 3D semantic parsing of large-scale indoor spaces, in: Proceedings of the IEEE Computer Society Conference on Computer Vision and Pattern Recognition. IEEE Computer Society, pp. 1534–1543. https://doi.org/10.1109/CVPR.2016.170

Bosch, M., Foster, K., Christie, G., Wang, S., Hager, G.D., Brown, M., 2018. Semantic Stereo for Incidental Satellite Images, in: ArXiv Preprint ArXiv:1811.08739.

Boulch, A., Guerry, J., Le Saux, B., Audebert, N., 2018. SnapNet: 3D point cloud semantic labeling with 2D deep segmentation networks. Computers and Graphics (Pergamon) 71, 189–198. https://doi.org/10.1016/j.cag.2017.11.010

Chehata, N., Guo, L., Mallet, C., 2009. Airborne Lidar Feature Selection for Urban Classification Using Random Forests, in: International Archives of Photogrammetry, Remote Sensing and Spatial Information Sciences.

Chen, Z., Gao, B., Devereux, B., 2017. State-of-the-Art: DTM Generation Using Airborne LIDAR Data. Sensors (Basel, Switzerland) 17. https://doi.org/10.3390/s17010150

Cooper, H.M., Chen, Q., Fletcher, C.H., Barbee, M.M., 2013. Assessing Vulnerability Due to Sea-Level Rise in Maui, Hawai'i Using Lidar Remote Sensing and GIS. Climatic Change 116, 547–563. https://doi.org/10.1007/s10584-012-0510-9

Dai, A., Chang, A.X., Savva, M., Halber, M., Funkhouser, T., Nießner, M., 2017. ScanNet: Richly-annotated 3D Reconstructions of Indoor Scenes, in: Proceedings of IEEE Conference on Computer Vision and Pattern Recognition, CVPR 2017. Institute of Electrical and Electronics Engineers Inc., pp. 2432–2443.

Feng, M., Zhang, L., Lin, X., Gilani, S.Z., Mian, A., 2020. Point attention network for semantic segmentation of 3D point clouds. Pattern Recognition 107, 107446. https://doi.org/10.1016/j.patcog.2020.107446

Fu, J., Liu, J., Tian, H., Li, Y., Bao, Y., Fang, Z., Lu, H., 2018. Dual Attention Network for Scene Segmentation. Proceedings of the IEEE Computer Society Conference on Computer Vision and Pattern Recognition 2019-June, 3141–3149.

Groh, F., Wieschollek, P., Lensch, H.P.A., 2019. Flex-Convolution: Million-Scale Point-Cloud Learning Beyond Grid-Worlds, in: Asian Conference on Computer Vision. Springer, pp. 105–122. https://doi.org/10.1007/978-3-030-20887-5_7

Hu, J., Shen, L., Sun, G., 2018. Squeeze-and-Excitation Networks, in: Proceedings of the IEEE Computer Society Conference on Computer Vision and Pattern Recognition. IEEE Computer Society, pp. 7132–7141. https://doi.org/10.1109/CVPR.2018.00745

Hu, X., Yuan, Y., 2016. Deep-Learning-Based Classification for DTM Extraction from ALS Point Cloud. Remote Sensing 8, 730. https://doi.org/10.3390/rs8090730

Huang, R., Xu, Y., Hong, D., Yao, W., Ghamisi, P., Stilla, U., 2020. Deep point embedding for urban classification using ALS point clouds: A new perspective from local to global. ISPRS Journal of Photogrammetry and Remote Sensing 163, 62–81. https://doi.org/10.1016/j.isprsjprs.2020.02.020

Jiang, M., Wu, Y., Zhao, T., Zhao, Z., Lu, C., 2018. PointSIFT: A SIFT-like Network Module for 3D Point Cloud Semantic Segmentation. arXiv.

Kalogerakis, E., Averkiou, M., Maji, S., Chaudhuri, S., 2017. 3D Shape Segmentation with Projective Convolutional Networks, in: Proceedings of the IEEE Conference on Computer Vision and Pattern Recognition. pp. 3779–3788.

Kohli, P., Ladický, L., Torr, P.H.S., 2009. Robust Higher Order Potentials for Enforcing Label Consistency. International Journal of Computer Vision 82, 302–324. https://doi.org/10.1007/s11263-008-0202-0

Landrieu, L., Obozinski, G., 2017. Cut Pursuit: Fast Algorithms to Learn Piecewise Constant Functions on General Weighted Graphs. SIAM Journal on Imaging Sciences 10, 1724–1766. https://doi.org/10.1137/17M1113436

Landrieu, L., Simonovsky, M., 2018. Large-scale Point Cloud Semantic Segmentation with Superpoint Graphs, in: Proceedings of the IEEE Conference on Computer Vision and Pattern Recognition. pp. 4558–4567.

Lemmen, C., van Oosterom, P., Bennett, R., 2015. The Land Administration Domain Model. Land Use Policy 49, 535–545. https://doi.org/10.1016/J.LANDUSEPOL.2015.01.014

Li, W., Wang, F.D., Xia, G.S., 2020a. A geometry-attentional network for ALS point cloud classification. ISPRS Journal of Photogrammetry and Remote Sensing 164, 26–40. https://doi.org/10.1016/j.isprsjprs.2020.03.016

Li, W., Wang, F.D., Xia, G.S., 2020b. A geometry-attentional network for ALS point cloud classification. ISPRS Journal of Photogrammetry and Remote Sensing 164, 26–40. https://doi.org/10.1016/j.isprsjprs.2020.03.016

Li, Y., Bu, R., Sun, M., Wu, W., Di, X., Chen, B., 2018. PointCNN: Convolution On X-Transformed Points, in: Advances in Neural Information Processing Systems. pp. 820–830.

Lin, C.-H., Chen, J.-Y., Su, P.-L., Chen, C.-H., 2014. Eigen-Feature Analysis of Weighted Covariance Matrices for LiDAR Point Cloud Classification. ISPRS Journal of Photogrammetry and Remote Sensing 94, 70–79. https://doi.org/10.1016/J.ISPRSJPRS.2014.04.016

Lin, Y., Vosselman, G., Cao, Y., Yang, M.Y., 2020. Active and incremental learning for semantic ALS point cloud segmentation. ISPRS Journal of Photogrammetry and Remote Sensing 169, 73–92. https://doi.org/10.1016/j.isprsjprs.2020.09.003

Lin, Y., Yang, M.Y., Nex, F., 2018. Semantic Building Façade Segmentation from Airborne Oblique Images, in: ISPRS TC II Mid-Term Symposium.

Lodha, S.K., Fitzpatrick, D.M., Helmbold, D.P., 2007. Aerial Lidar Data Classification using AdaBoost, in: Sixth International Conference on 3-D Digital Imaging and Modeling (3DIM 2007). IEEE, pp. 435–442. https://doi.org/10.1109/3DIM.2007.10

Lodha, S.K., Kreps, E.J., Helmbold, D.P., Fitzpatrick, D., 2006. Aerial LiDAR Data Classification Using Support Vector Machines (SVM), in: Third International Symposium on 3D Data Processing, Visualization, and Transmission (3DPVT'06). IEEE, pp. 567–574. https://doi.org/10.1109/3DPVT.2006.23

Mao, J., Wang, X., Li, H., 2019. Interpolated Convolutional Networks for 3D Point Cloud Understanding, in: Proceedings of the IEEE International Conference on Computer Vision. pp. 1578–1587.

Maturana, D., Scherer, S., 2015. VoxNet: A 3D Convolutional Neural Network for Real-Time Object Recognition, in: 2015 IEEE/RSJ International Conference on Intelligent Robots and Systems (IROS). pp. 922–928.

Meng, X., Currit, N., Wang, L., Yang, X., 2012. Detect Residential Buildings from Lidar and Aerial Photographs through Object-Oriented Land-Use Classification. Photogrammetric Engineering & Remote Sensing 78, 35–44. https://doi.org/10.14358/PERS.78.1.35

Murgante, B., Borruso, G., Lapucci, A., 2009. Geocomputation and Urban Planning, in: Geocomputation and Urban Planning. Springer Berlin Heidelberg, Berlin, Heidelberg, pp. 1–17. https://doi.org/10.1007/978-3-540-89930-3_1

Murtha, T., Golden, C., Cyphers, A., Klippel, A., Flohr, T., 2018. Beyond Inventory and Mapping: LIDAR, Landscape and Digital Landscape Architecture, in: Journal of Digital Landscape Architecture. pp. 249–259. https://doi.org/10.14627/537642027

Niemeyer, J., Rottensteiner, F., Soergel, U., 2014. Contextual classification of lidar data and building object detection in urban areas. ISPRS Journal of Photogrammetry and Remote Sensing 87, 152–165. https://doi.org/10.1016/J.ISPRSJPRS.2013.11.001

Niemeyer, J., Rottensteiner, F., Soergel, U., Heipke, C., 2016. Hierarchical Higher Order CRF for the Classification of Airborne Lidar Point Clouds in Urban Areas, in: ISPRS - International Archives of the Photogrammetry, Remote Sensing and Spatial Information Sciences. pp. 655–662. https://doi.org/10.5194/isprsarchives-XLI-B3-655-2016

Paszke, A., Gross, S., Massa, F., Lerer, A., Bradbury Google, J., Chanan, G., Killeen, T., Lin, Z., Gimelshein, N., Antiga, L., Desmaison, A., Xamla, A.K., Yang, E., Devito, Z., Raison Nabla, M., Tejani, A., Chilamkurthy, S., Ai, Q., Steiner, B., Facebook, L.F., Facebook, J.B., Chintala, S., 2019. PyTorch: An Imperative Style, High-Performance Deep Learning Library, in: Advances in Neural Information Processing Systems. pp. 8026–8037.

Qi, C.R., Su, H., Mo, K., Guibas, L.J., 2017a. PointNet: Deep Learning on Point Sets for 3D Classification and Segmentation, in: Proceedings of the IEEE Conference on Computer Vision and Pattern Recognition (CVPR). IEEE, p. 4. https://doi.org/10.1109/CVPR.2017.16

Qi, C.R., Yi, L., Su, H., Guibas, L.J., 2017b. PointNet++: Deep Hierarchical Feature Learning on Point Sets in a Metric Space, in: Advances in Neural Information Processing Systems. pp. 5099–5108.

Schmohl, S., Sörgel, U., 2019. Submanifold Sparse Convolutional Networks for Semantic Segmentation of Large-Scale ALS Point Clouds, in: ISPRS Annals of the Photogrammetry, Remote Sensing and Spatial Information Sciences. Copernicus GmbH, pp. 77–84. https://doi.org/10.5194/isprs-annals-IV-2-W5-77-2019

Shen, Y., Wu, L., Wang, Z., 2010. Identification of Inclined Buildings from Aerial LiDAR Data for Disaster Management, in: 2010 18th International Conference on Geoinformatics. IEEE, pp. 1–5. https://doi.org/10.1109/GEOINFORMATICS.2010.5567852

Simonovsky, M., Komodakis, N., 2017. Dynamic Edge-Conditioned Filters in Convolutional Neural Networks on Graphs. Proceedings - 30th IEEE Conference on Computer Vision and Pattern Recognition, CVPR 2017 2017-January, 29–38.

Tchapmi, L., Choy, C.B., Armeni, I., Gwak, J., Savarese, S., 2017. SEGCloud: Semantic Segmentation of 3D Point Clouds, in: 2017 International Conference on 3D Vision (3DV). IEEE, pp. 537–547.

Thomas, H., Qi, C.R., Deschaud, J.-E., Marcotegui, B., Goulette, F., Guibas, L.J., 2019. KPConv: Flexible and Deformable Convolution for Point Clouds, in: Proceedings of the IEEE International Conference on Computer Vision. pp. 6411–6420.

Tombari, F., Salti, S., Di Stefano, L., 2010. Unique signatures of histograms for local surface

description, in: European Conference on Computer Vision. Springer Verlag, pp. 356–369. https://doi.org/10.1007/978-3-642-15558-1_26

Vaswani, A., Brain, G., Shazeer, N., Parmar, N., Uszkoreit, J., Jones, L., Gomez, A.N., Kaiser, Ł., Polosukhin, I., 2017. Attention Is All You Need, in: Advances in Neural Information Processing Systems. pp. 5998–6008.

Vosselman, G., Coenen, M., Rottensteiner, F., 2017. Contextual segment-based classification of airborne laser scanner data. ISPRS Journal of Photogrammetry and Remote Sensing 128, 354–371. https://doi.org/10.1016/j.isprsjprs.2017.03.010

Vosselman, G., Maas, H.-G., 2010. Airborne and terrestrial laser scanning. CRC Press.

Wallace, L., Lucieer, A., Watson, C., Turner, D., Wallace, L., Lucieer, A., Watson, C., Turner, D., 2012. Development of a UAV-LiDAR System with Application to Forest Inventory. Remote Sensing 4, 1519–1543. https://doi.org/10.3390/rs4061519

Wang, X., Girshick, R., Gupta, A., He, K., 2018. Non-local Neural Networks, in: Proceedings of the IEEE Computer Society Conference on Computer Vision and Pattern Recognition. IEEE Computer Society, pp. 7794–7803. https://doi.org/10.1109/CVPR.2018.00813

Weinmann, M., Jutzi, B., Mallet, C., 2014. Semantic 3D Scene Interpretation: A Framework Combining Optimal Neighborhood Size Selection with Relevant Features, in: ISPRS Annals of the Photogrammetry, Remote Sensing and Spatial Information Sciences. pp. 181–188. https://doi.org/10.5194/isprsannals-II-3-181-2014

Weinmann, M., Jutzi, B., Mallet, C., 2013. Feature Relevance Assessment for the Semantic Interpretation of 3D Point Cloud Data Topology in Data Analysis View Project ANR Hiatus View Project Feature Relevance Assessment for the Semantic Interpretation of 3D Point Cloud Data. ISPRS Annals of the Photogrammetry, Remote Sensing and Spatial Information Sciences 5. https://doi.org/10.5194/isprsannals-II-5-W2-313-2013

Wen, C., Yang, L., Li, X., Peng, L., Chi, T., 2020. Directionally constrained fully convolutional neural network for airborne LiDAR point cloud classification. ISPRS Journal of Photogrammetry and Remote Sensing 162, 50–62. https://doi.org/10.1016/j.isprsjprs.2020.02.004

Winiwarter, L., Mandlburger, G., Schmohl, S., Pfeifer, N., 2019. Classification of ALS Point Clouds Using End-to-End Deep Learning. PFG - Journal of Photogrammetry, Remote Sensing and Geoinformation Science 87, 75–90. https://doi.org/10.1007/s41064-019-00073-0

Wu, W., Qi, Z., Fuxin, L., 2019. PointConv: Deep Convolutional Networks on 3D Point Clouds, in: Proceedings of the IEEE Conference on Computer Vision and Pattern Recognition. pp. 9621–9630.

Wu, Z., Song, S., Khosla, A., Fisher, Y., Zhang, L., Tang, X., Xiao, J., 2015. 3D ShapeNets: A Deep Representation for Volumetric Shapes, in: Proceedings of the IEEE Conference on Computer Vision and Pattern Recognition. pp. 1912–1920.

Xiong, X., Munoz, D., Bagnell, J.A., Hebert, M., 2011. 3-D scene analysis via sequenced predictions over points and regions, in: Proceedings - IEEE International Conference on Robotics and Automation. pp. 2609–2616. https://doi.org/10.1109/ICRA.2011.5980125

Xu, S., Vosselman, G., Oude Elberink, S., 2014. Multiple-Entity Based Classification of Airborne Laser Scanning Data in Urban Areas. ISPRS Journal of Photogrammetry and Remote Sensing 88, 1–15. https://doi.org/10.1016/J.ISPRSJPRS.2013.11.008

Xu, Y., Fan, T., Xu, M., Zeng, L., Qiao, Y., 2018. SpiderCNN: Deep Learning on Point Sets

with Parameterized Convolutional Filters, in: Proceedings of the European Conference on Computer Vision (ECCV). pp. 87–102.

Yang, Z., Jiang, W., Xu, B., Zhu, Q., Jiang, S., Huang, W., Yang, Z., Jiang, W., Xu, B., Zhu, Q., Jiang, S., Huang, W., 2017. A Convolutional Neural Network-Based 3D Semantic Labeling Method for ALS Point Clouds. Remote Sensing 9, 936. https://doi.org/10.3390/rs9090936

Yang, Z., Tan, B., Pei, H., Jiang, W., 2018. Segmentation and Multi-Scale Convolutional Neural Network-Based Classification of Airborne Laser Scanner Data. Sensors 18, 3347. https://doi.org/10.3390/s18103347

Yousefhussien, M., Kelbe, D.J., Ientilucci, E.J., Salvaggio, C., 2018. A multi-scale fully convolutional network for semantic labeling of 3D point clouds. ISPRS Journal of Photogrammetry and Remote Sensing 143, 191–204. https://doi.org/10.1016/j.isprsjprs.2018.03.018

Zhao, R., Pang, M., Wang, J., 2018. Classifying airborne LiDAR point clouds via deep features learned by a multi-scale convolutional neural network. International Journal of Geographical Information Science 32, 960–979. https://doi.org/10.1080/13658816.2018.1431840

Zheng, S., Jayasumana, S., Romera-Paredes, B., Vineet, V., Su, Z., Du, D., Huang, C., Torr, P.H.S., 2015. Conditional Random Fields as Recurrent Neural Networks, in: Proceedings of the IEEE International Conference on Computer Vision. pp. 1529–1537.